\documentclass[
    conference,
    10pt,
    ]{IEEEtran}
\usepackage[utf8]{inputenc}
\usepackage[T1]{fontenc}

\usepackage[ruled,vlined,linesnumbered]{algorithm2e}
\usepackage{amsmath}

\usepackage{amsthm}
\usepackage{graphicx}
\usepackage{amssymb}
\usepackage{ragged2e}
\usepackage{flushend}
\usepackage{balance}
\makeatletter
\@namedef{ver@lineno.sty}{9999/12/31}
\@namedef{opt@lineno.sty}{}
\makeatother
\usepackage[frozencache=true,cachedir=minted-output]{minted}  
\usepackage{siunitx}
\usepackage{xcolor}

\theoremstyle{definition}
\newtheorem{definition}{Definition}

\makeatletter
\let\MYcaption\@makecaption
\makeatother
\usepackage[font=footnotesize]{subcaption}
\makeatletter
\let\@makecaption\MYcaption
\makeatother

\usepackage{hyperref}
\usepackage[capitalise]{cleveref}

\title{Anomaly Detection with \\ Neural Parsers That Never Reject}
\author{%
\IEEEauthorblockN{Alexander Grushin}%
    \IEEEauthorblockA{\small Galois, Inc.\\
        \texttt{agrushin@galois.com}}
\and
\IEEEauthorblockN{Walt Woods}%
    \IEEEauthorblockA{\small Galois, Inc.\\
        \texttt{waltw@galois.com}}
}
\date{February 2021}

\begin{document}



\date{}
\maketitle
\thispagestyle{plain}
\pagestyle{plain}

\begin{abstract} 
    Reinforcement learning has recently shown promise as a technique for training an artificial neural network to parse sentences in some unknown format, through a body of work known as RL-GRIT.  A key aspect of the RL-GRIT approach is that rather than explicitly inferring a grammar that describes the format, the neural network learns to perform various parsing actions (such as merging two tokens) over a corpus of sentences, with the goal of maximizing the estimated frequency of the resulting parse structures.  This can allow the learning process to more easily explore different action choices, since a given choice may change the optimality of the parse (as expressed by the total reward), but will not result in the failure to parse a sentence.  However, this also presents a limitation: because the trained neural network can successfully parse any sentence, it cannot be directly used to identify sentences that deviate from the format of the training sentences, i.e., that are anomalous.  In this paper, we address this limitation by presenting procedures for extracting production rules from the neural network, and for using these rules to determine whether a given sentence is nominal or anomalous. When a sentence is anomalous, an attempt is made to identify the location of the anomaly. We empirically demonstrate that our approach is capable of grammatical inference and anomaly detection for both non-regular formats and those containing regions of high randomness/entropy. While a format with high randomness typically requires large sets of production rules, we propose a two pass grammatical inference method to generate parsimonious rule sets for such formats. By further improving parser learning, and leveraging the presented rule extraction and anomaly detection algorithms, one might begin to understand common errors, either benign or malicious, in practical formats.

        
    
    
    
\end{abstract}



\section{Introduction}\label{sec:intro}



Grammatical inference is desirable in many domains, limited not only to {\em Natural Language Processing} (NLP), but also the processing of data that adheres to unknown or underspecified formats. For example, one use case arises when enterprise systems seek security by pre-processing data, before passing it to any application.  To ensure compatibility with an application, it may be necessary to infer the grammar for the format that the application accepts. With large, well-adopted formats, such as {\em Portable Document Format} (PDF), there exists an ecosystem of programs (parsers) for reading and writing files in the format. These parsers do not always adhere to the specification's original intent, and sometimes add features or contain bugs, resulting in a modified, de facto specification. Understanding these modifications to design an effective pre-filter requires grammatical inference algorithms, which can represent and capture the complicated data structures often found in data formats.

Given a set of example input {\em sentences} in some unknown format, an effective grammatical inference algorithm must generate a grammar that is general enough to correctly and completely describe this format, i.e., that will parse all sentences in that format, but no other sentences.  
In order to improve the generality of grammatical inference for non-trivial formats, recent work has often leveraged advances in {\em Machine Learning} (ML), particularly, via artificial neural networks.  For example, methods have been developed for extracting a grammar from a recurrent neural network, which is trained to estimate the probability of a given token in a sentence, given other tokens \cite{yellin2021synth,barbot2021extracting}; the neural network is treated as a parser that accepts a sentence if the probabilities are sufficiently high, and rejects it otherwise. Experiments have shown that the approach can successfully infer certain context-free grammars, e.g., for Dyck languages.  However, while the grammar extraction approach was shown to be robust to neural network errors, due to imperfect training \cite{yellin2021synth}, it is not presently known whether it would be effective at capturing formats where valid sentences can contain regions with a high degree of randomness/entropy, interspersed with low-entropy regions -- e.g., the mixing of control bytes, which denote structure within a format, with user data contained in the payload of a format.  In a high-entropy region, the probability of one token (e.g., \mintinline[breaklines]{text}|a| at the end of the randomly-generated sentence \mintinline[breaklines]{text}|PxdGmfn3ea|) is approximately the same as the probability of any other (e.g., \mintinline[breaklines]{text}|b| at the end of \mintinline[breaklines]{text}|PxdGmfn3eb|); individual token probabilities are thus all low, which can potentially cause a valid sentence to be rejected, unless the trained parser is highly accurate.  (In this paper, we use the term {\em token} to refer to a single character in a sentence, though a tokenizer or lexer could be used to extend our algorithm to multi-character tokens.)

\begin{table*}[]
\centering
\caption{Comparison to related work}
\label{table:introduction:comparison}
{\footnotesize
\begin{tabular}{|l|c|c|c|}
\hline
\textbf{Approach}         & \textbf{Data Type Recurrency} & \textbf{High-Entropy Regions} & \textbf{Anomaly Detection} \\ \hline
Recurrent Neural Network and Grammar Extraction \cite{yellin2021synth,barbot2021extracting} & \checkmark & {\em ?} & \checkmark \\ \hline
Constituency Parsing via Autoencoders or Transformers \cite{drozdov2019diora,drozdov2020sdiora,wang2019treetransformer} & {\em X} & \checkmark & {\em X} \\ \hline
Reinforcement Learning \cite{cowger2020icarus,woods2021rlgrit} & \checkmark & \checkmark & {\em X} \\ \hline
Reinforcement Learning and Grammar Extraction (this study) & \checkmark & \checkmark & \checkmark \\ \hline
\end{tabular}

\bigskip
\begin{justify}
Each row corresponds to a particular grammatical inference approach, while columns indicate a specific feature; for a given approach-feature combination, a \checkmark symbol indicates that the approach supports the feature; an {\em X} symbol indicates that the approach does not directly support the feature; a {\em ?} symbol indicates that it has not been established (to our knowledge) whether the approach supports the feature.
\end{justify}
}
\end{table*}

Other work applied deep recursive autoencoders \cite{drozdov2019diora,drozdov2020sdiora} and transformers \cite{wang2019treetransformer} to perform {\em constituency parsing}, where adjacent, related {\em atoms} within a sentence are concatenated to form larger atoms.  As defined here, an atom can be a single {\em token} in a parsed sentence, or can consist of multiple such tokens that were merged together during previous parsing steps; e.g., \mintinline[breaklines]{text}|'if'| would be an atom formed from the atoms \mintinline[breaklines]{text}|'i'| and \mintinline[breaklines]{text}|'f'|. In our notation, we use single quotes to indicate that a sequence of tokens should be treated as an atom.  Such approaches can potentially produce a {\em parse tree} for any sentence, including one with high-entropy regions, and can thus potentially learn the low-entropy ``subformat'' of a format.  In such a parse tree, nodes represent atoms, while parent-child relationships denote merges; for example, \mintinline[breaklines]{text}|'i'| and \mintinline[breaklines]{text}|'f'| would be the child nodes of \mintinline[breaklines]{text}|'if'|.  However, these approaches do not have the appropriate expressiveness for {\em data type recurrency}.  For example, in the {\em JavaScript Object Notation} (JSON) format, a nested object may appear as follows: \mintinline[breaklines]{text}|{ "a": { "b": { "c": "d" } } }|.  Here, both the \mintinline[breaklines]{text}|"a"| and the \mintinline[breaklines]{text}|"b"| keys map to object values, but the above approaches might represent the first object as the atom \mintinline[breaklines]{text}|'{ "b": { "c": "d" } }'|, and the second one as the atom \mintinline[breaklines]{text}|'{"c" : "d" }'|, i.e., as different data types.

A third recent body of work combines constituency parsing with {\em Reinforcement Learning} (RL), as the basis for a novel, flexible grammatical inference scaffolding called RL-GRIT \cite{cowger2020icarus,woods2021rlgrit}. This approach attempts to handle high-entropy regions while maintaining expressiveness. The goal of RL is to find a {\em policy}, defined as a mapping of {\em observations} to {\em actions}, which attempts to maximize some {\em reward}. In RL-GRIT, an observation is the current set of atoms, while an action involves the merging of two atoms into a new higher-level atom (parsing is thus performed in a bottom-up fashion).  Importantly, to overcome the aforementioned limitations associated with simply concatenating atoms, there are special merge action types that enable data type recurrency: when merging two atoms, the parser has the option of replacing one of the atoms with a special {\em subgrammar token}, which can act as a wildcard that can match multiple atoms, or removing one of the atoms entirely, which allows multiple atoms to be recursively collapsed into a single atom, achieving an effect similar to that of the Kleene star.  The reward is roughly based on the estimated frequency of the atoms generated when performing parsing actions on a corpus of sentences. The learned policy, represented as a neural network, can then be used as a parser, which can be applied to generate a parse tree for any new sentence.  This RL-based approach learns a parser directly, rather than first learning a grammar.  The primary reason for this is that the learning process benefits from a search space with progressive improvements in the measured reward, rather than the ``all-or-nothing'' design of traditional grammars, which either accept or reject a sentence.

While the above approach was shown to be promising at automatically generating parsers for formats with data type recurrency and high-entropy regions \cite{cowger2020icarus,woods2021rlgrit}, it also has limitations.  In particular, because the parser is always expected to take an action at each step of the parsing process, it will produce a parse tree for {\em any} input sentence, even if the format of this sentence is entirely different from the format of sentences that were used for training. Thus, unlike traditional parsers, which will fail upon encountering a deviation from the grammar, the RL-based parser cannot be used directly to determine whether some sentence is {\em nominal} (i.e., valid) or {\em anomalous} (invalid) relative to the training data.  Additionally, as is often the case with neural network-based approaches, there is a lack of {\em explainability}: the parser has learned some grammar, but it does not provide the production rules for that grammar.

\begin{figure*}[t]
    \centering
    \includegraphics[clip, trim={0in 3.70in 2.38in 0in}, width=\linewidth]{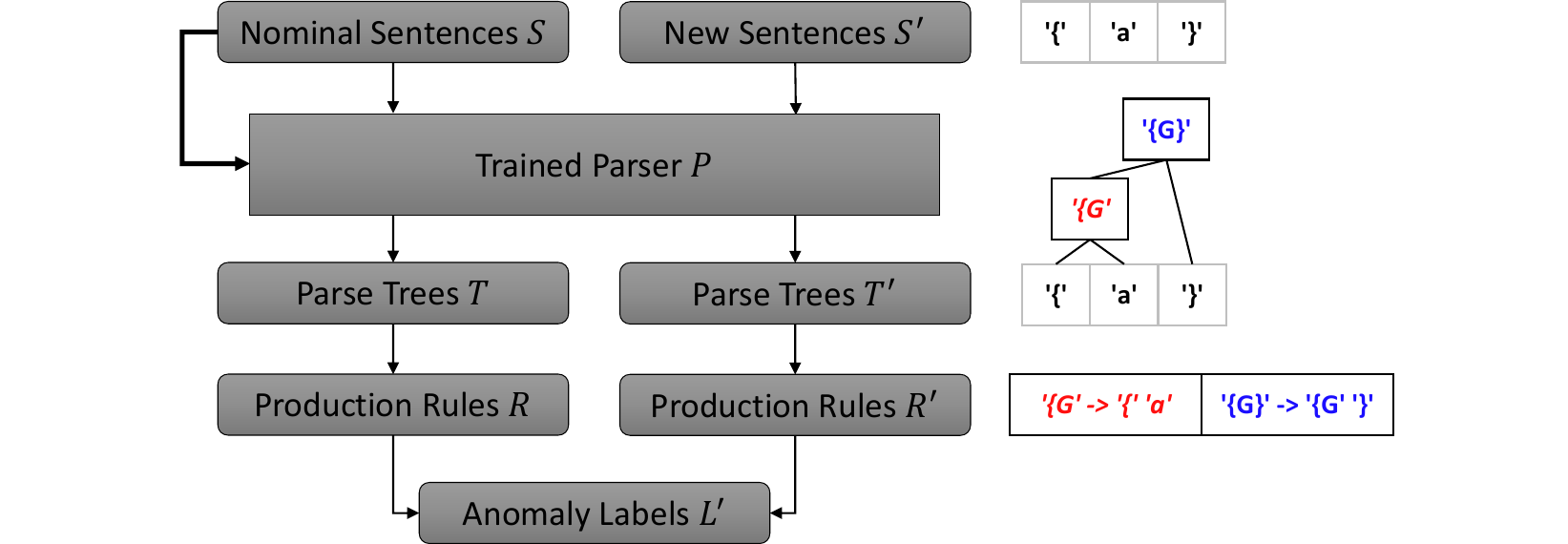}
    \caption{An outline of the approach that is presented in this paper (left), with very simple examples (right) of a sentence, a parse tree, and a rule.  The thick black arrow (left) represents the training of the parser on nominal sentences; thin black arrows represent data flows.  Details are provided in the text.}
    \label{fig:proposed:parser:overview}
\end{figure*}

In this paper, we extend the RL-based grammatical inference approach by introducing a simple {\em rule extraction} technique, which analyzes parse trees output by the trained parser and extracts a set of production rules describing the format.  We design a representation for the production rules, initially based directly on the actions that are taken by the parser. Subsequently, we identify and mitigate limitations of this approach to increase the expressive power of the representation.  As the merges are binary, the resulting representation somewhat resembles the  Chomsky Normal Form \cite{chomsky1959normalform}, though proving equivalence (or lack thereof) is a subject for future work.  We also provide a technique for {\em anomaly detection}, which uses the resulting representation to determine whether some new sentence adheres to or deviates from the format. To achieve this, our technique checks whether the new sentence is parsed using some of the same production rules that were extracted from sentences that are known to be nominal, or whether certain {\em unexpected} (new) production rules are applied.  In the latter case, the technique also attempts to find the location of the anomaly within the sentence, by identifying regions of the sentence that are parsed by the unexpected rules.  Finally, we further extend rule extraction and anomaly detection to formats where sentences can contain regions with a high degree of entropy.  When such a region appears in some new sentence, there is a risk that it will be parsed by some unexpected rules, and will thus be labeled as anomalous, even if it is not anomalous in the context of the grammar being learned, as it represents a subformat which is not integral to the containing format's structure. We propose to deal with this by applying our approach in two passes: once anomaly detection (potentially, with some modifications) has identified high-entropy regions, they are removed, yielding a simpler set of sentences; then, the ``train a parser, extract rules, detect anomalies'' pipeline is applied a second time, to these simplified sentences. The resulting rules capture the low-entropy subformat of the original format, and allow potential problems to be identified in this subformat, without being confounded with high-entropy regions.  As we outline in \cref{table:introduction:comparison}, to our knowledge, our approach is the first to demonstrate anomaly detection in formats with high-entropy regions, while also being capable of handling formats with data type recurrency.

\section{Approach}\label{sec:approach}

Our approach is outlined in \cref{fig:proposed:parser:overview}.  Given some set of example sentences that are nominal (i.e., valid, according to some unknown format), we use RL (specifically, RL-GRIT) to train a parser $P$ on some subset of these sentences.  We apply the trained parser $P$ to another subset of these sentences, in order to generate parse trees $T$, and to then extract a set of rules $R$ (we use disjoint subsets of $S$ for training vs. rule extraction, though this need not be the case).  Given some new sentences $S^{\prime}$ (whose format is unknown), we generate parse trees $T^{\prime}$ for these sentences as well, and extract rules $R^{\prime}$.  By comparing the rules $R$ and $R^{\prime}$, determinations can be made regarding whether each sentence in $S^{\prime}$ is nominal or anomalous, and in the latter case, where the anomalies might potentially exist.  We elaborate upon these steps in the following subsections.


\subsection{Learned actions and production rules}\label{sec:proposed:actions}

RL is a powerful ML paradigm for searching spaces of policies (i.e., observation-action mappings). That is, RL seeks to find a mapping from environmental observations to an action sequence, which optimizes some reward. Typically, this is applied to situations such as a robot navigating a maze based on sensory inputs. However, RL provides a generic framework for efficiently enumerating policies consisting of discrete actions, making it one of the most flexible paradigms for searching any decision making process. Success on large search spaces has been achieved when neural networks are used to track expected rewards \cite{mnih2013atari}. Recent work on applying RL to the grammatical inference problem was initially presented at LangSec 2020 \cite{cowger2020icarus}, and the idea has since been significantly expanded and refined in \cite{woods2021rlgrit}, presented at LangSec 2021; we leverage the RL-GRIT algorithm from the latter paper in the present work.  General RL background can be found in a number of papers, such as \cite{mnih2013atari,watkins1992qlearning,mnih2016a2c}.

\begin{figure*}[t]
    \centering
    \includegraphics[clip, trim={0in 2.95in 2.38in 0in}, width=\linewidth]{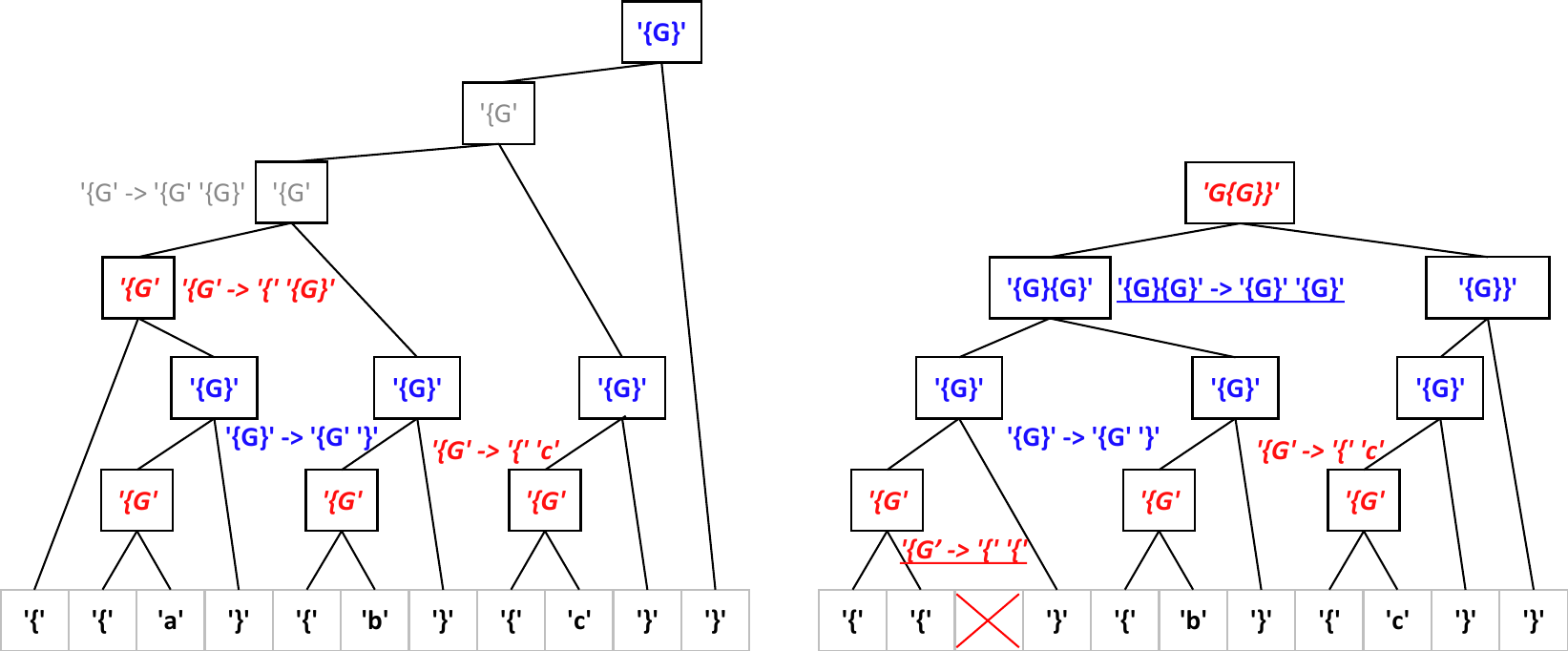}
    \caption{Example nominal (left) and anomalous (right) Simple-JSON sentences, with parse trees; the leaves of the parse tree correspond to single-character tokens in the sentence.  Each node of the parse tree is labeled with the atom that it represents. Some nodes are also labeled with the production rule that generated the atom (i.e., that has this atom on the left side).  Blue, bold rules (and generated atoms) correspond to regular merges; grey rules correspond to anchored merges, and red, bold, italicized rules correspond to subgrammar merges.  Unexpected rules are underlined.}
    \label{fig:proposed:parser:trees}
\end{figure*}

Here, we focus specifically on the actions that the artificial neural network-based parser learns to perform.  At every step of the parse, an action involves selecting two adjacent atoms, and merging them into a new atom.  In the following, let us assume that the two atoms are \mintinline[breaklines]{text}|'a'| and \mintinline[breaklines]{text}|'b'|.  There are three types of merges; the most straightforward of these simply concatenates the two atoms; the result of the merge is the atom \mintinline[breaklines]{text}|'ab'|; this can be expressed via the production rule \mintinline[breaklines]{text}|'ab' -> 'a' 'b'|.  The {\em anchored merge} takes two atoms, \mintinline[breaklines]{text}|'a'| and \mintinline[breaklines]{text}|'b'|, and merges them into the atom \mintinline[breaklines]{text}|'a'| or \mintinline[breaklines]{text}|'b'| (essentially, deleting the other atom), depending on whether the anchored merge is left- or right-biased; these merges are described, respectively, as: \mintinline[breaklines]{text}|'a' -> 'a' 'b'| and \mintinline[breaklines]{text}|'b' -> 'a' 'b'|.  Finally, in the {\em subgrammar merge}, one of the two merged atoms is replaced with a special, wildcard-like token \mintinline[breaklines]{text}|'G'| not occurring in the input language.  Like the anchored merges, subgrammar merges can be left- or right-biased: \mintinline[breaklines]{text}|'aG' -> 'a' 'b'| and \mintinline[breaklines]{text}|'Gb' -> 'a' 'b'|.  As another example, in \cref{fig:proposed:parser:overview}, the left-biased subgrammar merge combines the atoms \mintinline[breaklines]{text}|'{'| and \mintinline[breaklines]{text}|'a'| into the atom \mintinline[breaklines]{text}|'{G'|, via the production rule  \mintinline[breaklines]{text}|'{G' -> '{' 'a'|.  Notably, the anchored merge is designed to fill a similar role to that of the Kleene star \mintinline[breaklines]{text}|*| (e.g., it can be applied multiple times to parse a sentence such as \mintinline[breaklines]{text}|abbb| to produce \mintinline[breaklines]{text}|abb|, then \mintinline[breaklines]{text}|ab|, and then \mintinline[breaklines]{text}|a|), and the subgrammar merge is designed to fill a similar role to that of the alternation operator \mintinline[breaklines]{text}|||.

We formally define the merge actions (and their corresponding rules) as follows:

\begin{definition}
Let a {\em sentence} $S_{i} = (c_{1}, c_{2}, ..., {c_n}) \in S$ be a sequence (list) of {\em tokens} $c_{j}$.  Let an {\em atom} $a = (A, b, e)$ be a tuple, where $A$ is a sequence of tokens that corresponds to (but is not necessarily the same as) the subsequence of tokens $(c_{b}, ..., c_{e})$ in $S_{i}$, with $1 \leq b \leq e \leq n$.  For each token $c_{j}$, there is a corresponding atom $((c_{j}), j, j)$; such an atom is also known as a {\em leaf}.  Additionally, an atom $a = m(a_{L},a_{R})$ may be the result of a {\em merge} $m$ of two adjacent atoms $a_{L} = (A_{L}, b_{L}, k)$ and $a_{R} = (A_{R}, k+1, e_{R})$; here, $a_{L}$ and $a_{R}$ are the {\em left child} and {\em right child} of $a$, respectively.  The merge is described via the {\em production rule} $a \rightarrow a_{L}\:a_{R}$.  The following merge types are considered:

\begin{enumerate}
    \item If $m$ is a {\em regular merge}, then $a = (A_{L} + A_{R}, b_{L}, e_{R})$, where the $+$ operator denotes the concatenation of two sequences.
    \item If $m$ is a {\em left-biased anchored merge}, then $a = (A_{L}, b_{L}, e_{R})$
    \item If $m$ is a {\em right-biased anchored merge}, then $a = (A_{R}, b_{L}, e_{R})$.
    \item If $m$ is a {\em left-biased subgrammar merge}, then $a = (A_{L} + (G), b_{L}, e_{R})$, where $G$ is the {\em subgrammar token} (that does not appear in any sentence in $S$), and (G) is a sequence consisting of just the subgrammar token $G$.
    \item If $m$ is a {\em right-biased subgrammar merge}, then $a = ((G) + A_{R}, b_{L}, e_{R})$.
\end{enumerate}
\end{definition}

We note that throughout the text, we often informally refer to $A$ as an ``atom'', though formally, an atom $a = (A, b, e)$ also includes the indices $b$ and $e$.

\subsection{Production rule extraction}\label{sec:proposed:extraction}

\begin{figure*}[t]
    \centering
    \includegraphics[clip, trim={0in 4.07in 2.38in 0in}, width=\linewidth]{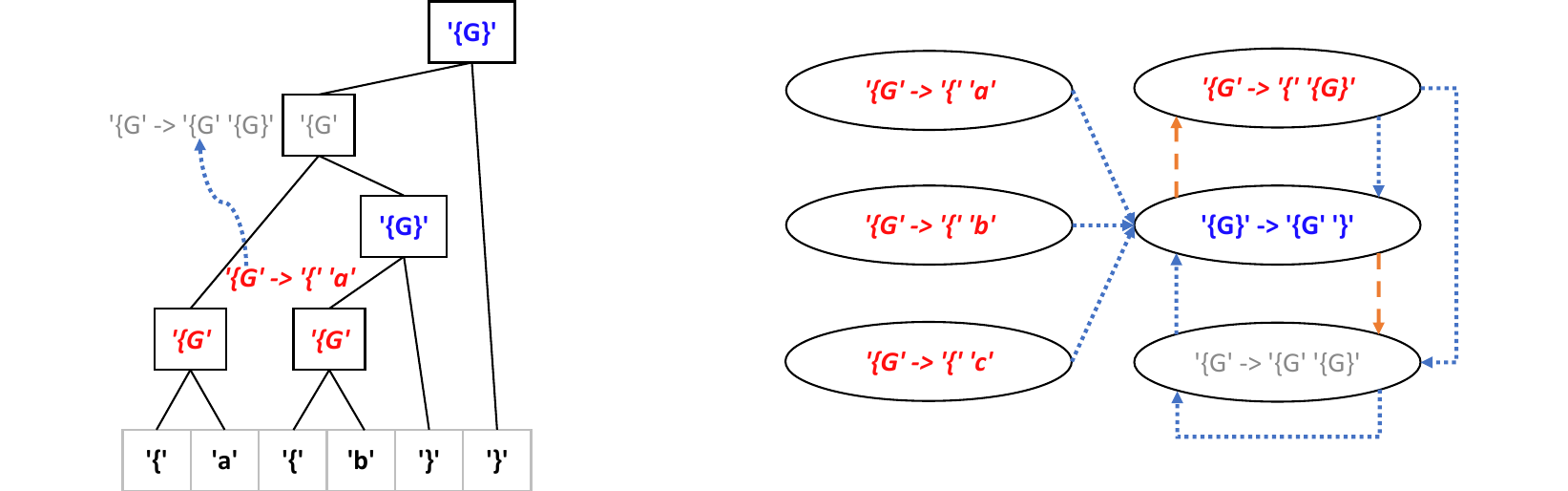}
    \caption{An example sentence that that does not adhere to the Simple-JSON format, but that would be labeled as nominal without precedence constraints (left); the blue, dotted arrow indicates that the left side of one production rule is fed into the right side of another production rule.  We illustrate some of the precedence constraints between production rules that parse valid/nominal Simple-JSON sentences (right); each node is labeled with a rule; a blue, dotted arrow indicates that the left side of one production rule becomes the {\em first} atom on the right side of another production rule; an orange, dashed arrow indicates that one rule produces the {\em second} atom that is used by another rule.}
    \label{fig:proposed:parser:precedence}
\end{figure*}

For a sentence consisting of $n$ tokens, there are $n-1$ possible merges as a first action, since all such actions from \cite{woods2021rlgrit} involve two adjacent atoms. For the same reason, the resulting tree from a sentence with $n$ tokens will be a binary tree with $n-1$ non-leaf nodes.  We call this binary tree a {\em parse tree}, and formally define it as follows:

\begin{definition}
Define the {\em descendants} of an atom $a$ to be an empty set if $a$ is a leaf.  Otherwise, let the descendants of $a$ be the union of the descendants of its left child and the descendants of its right child.  A {\em parse tree} $T_{i}$ for a sentence $S_{i}$ is an atom $a = (A, 1, n)$ (where $n$ is the length of $S_{i}$), and all descendants of $a$.  In the context of a parse tree, atoms are referred to as {\em nodes}.
\end{definition}

Each non-leaf node in the parse tree will be the result of some merge, and thus has an associated production rule that describes this merge.  Thus, by examining the parse tree, we can easily obtain the multiset of production rules that were used to parse the sentence (we keep track of the number of times that a given rule occurs).  Given a list of parse trees $T$, the multiset of rules $R$ can be generated via \cref{alg:extraction}.  The outer loop of the algorithm iterates over different parse trees that were obtained for different example sentences, since these trees may yield different subsets of rules, which are combined into a multiset $R$, to obtain a comprehensive view of the grammar that the parser has learned.

As an example, consider the sentence \mintinline[breaklines]{text}|{{a}{b}{c}}|, and its parse tree, illustrated on the left side of \cref{fig:proposed:parser:trees}; some of the non-leaf nodes are labeled with the associated rules (additional extracted rules, not shown in the figure, are  \mintinline[breaklines]{text}|'{G' -> '{' 'a'| and  \mintinline[breaklines]{text}|'{G' -> '{' 'b'|).  In this particular case, even though the production rules were extracted from a single parse tree, they happen to cover all sentences described by the Dyck-like grammar \mintinline[breaklines]{text}`S -> '{' ('a' | 'b' | 'c' | S+) '}'`, referred to as the {\em Simple-JSON} grammar \cite{cowger2020icarus,woods2021rlgrit}; in particular, the \mintinline[breaklines]{text}|'{G}'| atom (appearing on the left side of the production rule \mintinline[breaklines]{text}|'{G}' -> '{G' '}'|) approximately corresponds to \mintinline[breaklines]{text}|S|.  However, while the rules will successfully parse any valid Simple-JSON sentence, it is not yet clear whether they will fail to parse any sentence that is not in the Simple-JSON format, i.e., that is anomalous. We will address this question in the following two subsections.

\begin{algorithm}
\begin{enumerate}
    \item Input the list of parse trees $T$.
    \item Initialize the multiset of rules $R = \emptyset$.
    \item For each $T_{i} \in T$:
    \begin{enumerate}
        \item For each non-leaf node $a \in T_{i}$:
        \begin{enumerate}
            \item Let $a_L$ and $a_R$ be the left and right child of $a$, respectively.
            \item Define the rule $R_{j} = a \rightarrow a_L\:a_R$.
            \item Add $R_{j}$ to $R$.
        \end{enumerate}
    \end{enumerate}
    \item Output $R$.
\end{enumerate}
\caption{Rule Extraction} \label{alg:extraction}
\end{algorithm}


\subsection{Anomaly detection}\label{sec:proposed:detection}

During both parser training and rule extraction, our assumption was that all sentences are valid examples of a given format.  Now, suppose that we are presented with some new sentence $s^{\prime} \in S^{\prime}$, whose format is unknown.  Our goal is to use the trained parser $P$ and the extracted rules $R$ to determine whether $s^{\prime}$ is valid in the given format, or whether it deviates from it.  This determination can be made by applying the parser to $s^{\prime}$, and obtaining a multiset of rules $R^{\prime}_{s^{\prime}}$ for that sentence.  Then, anomaly detection can be performed via a very simple procedure, given as \cref{alg:detection}.  Specifically, we determine whether some of the rules $R^{\prime}_{s^{\prime}}$ that were extracted when parsing $s^{\prime}$ are {\em unexpected}, in that they were never used to parse valid/nominal sentences.

\begin{algorithm}
\begin{enumerate}
    \item Input the multisets of rules $R$ and $R^{\prime}_{s^{\prime}}$.
    \item If $R^{\prime}_{s^{\prime}} \subseteq R$:
    \begin{enumerate}
        \item Output a ``nominal'' label.
    \end{enumerate}
    \item Otherwise:
    \begin{enumerate}
        \item Output an ``anomalous'' label.
    \end{enumerate}
\end{enumerate}
\caption{Anomaly Detection} \label{alg:detection}
\end{algorithm}

As an example, consider the right side of \cref{fig:proposed:parser:trees}, where we apply the parser $P$ to a sentence $s^{\prime}$ that is a corrupted version of the sentence on the left side, with the \mintinline[breaklines]{text}|'a'| token deleted.  The analysis of the parse tree reveals a number of unexpected rules, such as \mintinline[breaklines]{text}|'{G}{G}' -> '{G}' '{G}'|, and the anomaly is thus correctly detected.

\subsection{Enhancing the representation}\label{sec:proposed:enhancements}

There are, however, situations where our approach may fail to detect that a sentence is anomalous.  First, consider a sentence such as \mintinline[breaklines]{text}|{a|. While it is anomalous (due to a missing bracket), it is successfully parsed via the rule \mintinline[breaklines]{text}|'{G' -> '{' 'a'|, which is expected (i.e., in $R$); thus, the sentence is labeled as nominal.  This shortcoming exists because our representation lacks the notion of a {\em start symbol}, which is a nonterminal symbol that must correspond to the entire sentence that is being parsed.  In our grammar representation, we now introduce a similar notion of a {\em start rule}, which is any rule that appears at the root of the parse trees that are produced by the parser $P$.  In particular, when rules are extracted from a parse tree, any rule at the root is labeled as a start rule; in our notation, we denote a start rule by enclosing its left side with hyphens.  For example, in the left (nominal) parse tree of \cref{fig:proposed:parser:trees}, the root of the tree is represented by the rule \mintinline[breaklines]{text}|'{G}' -> '{G' '}'|, so we write it as \mintinline[breaklines]{text}|-'{G}'- -> '{G' '}'|.
Given this enhancement, when performing anomaly detection on the aforementioned example \mintinline[breaklines]{text}|'{a'|, the resulting parse tree contains a single rule \mintinline[breaklines]{text}|-'{G'- -> '{' 'a'| at its root.  As the original set of rules $R$ does not contain the rule \mintinline[breaklines]{text}|-'{G'- -> '{' 'a'|, even though it does contain the rule \mintinline[breaklines]{text}|'{G' -> '{' 'a'|, the sentence \mintinline[breaklines]{text}|'{a'| is now be correctly labeled as anomalous.


Now, consider the sentence \mintinline[breaklines]{text}|{a{b}}|, and its parse tree, shown on the left side of \cref{fig:proposed:parser:precedence}.  This sentence is anomalous, due to an extra letter between the first two brackets.  However, the anomaly is not detected, because none of the extracted rules (including the start rule) are unexpected. To detect this sort of anomaly, we must consider not only {\em which} rules are applied to parse it, but also, {\em in what order}.  In particular, the goal is to determine exactly which production rules (or terminal symbols) might be used as inputs for which subsequent production rules.  For example, is it acceptable for the output of the rule \mintinline[breaklines]{text}|'{G' -> '{' 'a'| to be used as an input to the rule \mintinline[breaklines]{text}|'{G' -> '{G' '{G}'|, as indicated by the blue arrow in the figure?  To allow such questions to be answered, when the rule extraction procedure analyzes parse trees $T$ derived from nominal sentences, it considers not only rules, but also, {\em precedence constraints}, which we define as follows:

\begin{definition}
A {\em precedence constraint} takes the form $(w \succ u \land v)$, where $w$ is a rule, while $u$ and $v$ may be either a rule or a token.  The precedence constraint indicates that there is some parse tree $T_{i} \in T$ that contains a non-leaf node with a rule $w$, where the left child has a rule $u$ (if it is a non-leaf node) or a token $u$ (if it is a leaf node), while the right child has a rule $v$ (if it is a non-leaf node) or a token $v$ (if it is a leaf node).
\end{definition}

Together, the rules $R$ and the precedence constraints $C$ can be illustrated as a graph, with the former represented via nodes, and the latter captured as pairs of edges.  If we ignore, for clarity, the distinction between start rules and other rules, and only consider constraints where both $u$ and $v$ are rules, rather than tokens, then for Simple-JSON, the graph is shown on the right side of \cref{fig:proposed:parser:precedence}.  Now, when performing anomaly detection, we determine not only if the rules $R^{\prime}_{s^{\prime}}$ extracted from a new input sentence $s^{\prime} \in S^{\prime}$ exist in $R$, but also, whether or not the precedence constraints $C^{\prime}_{s^{\prime}}$ for that sentence exist in $C$; $s^{\prime}$ is labeled as nominal if and only if $R^{\prime}_{s^{\prime}} \subseteq R$ {\em and} $C^{\prime}_{s^{\prime}} \subseteq C$.  Returning to the example in \cref{fig:proposed:parser:precedence}, we observe that one of the extracted precedence constraints is as follows: \mintinline[breaklines]{text}|( '{G' -> '{G' '{G}' > '{G' -> '{' 'a' ^ '{G}' -> '{G' '}' )|.  This constraint does not appear in the original set of constraints in \cref{fig:proposed:parser:precedence} (there is no edge from \mintinline[breaklines]{text}|'{G' -> '{' 'a'| to \mintinline[breaklines]{text}|'{G' -> '{G' '{G}'|); thus, the example is now correctly labeled as anomalous.

\subsection{Anomaly localization}\label{sec:proposed:localization}

It is of practical interest to not only determine whether a sentence is anomalous, but also to localize the anomalous token(s) within the sentence.  We attempt to achieve this by determining which of the sentence's tokens are ``covered'' by the unexpected rules, or by rules $w$ with unexpected precedence constraints $(w \succ u \land v) \notin C$.  A token is {\em covered} by a rule if this rule applies at some node in the parse tree for the sentence, and a child of that node is a leaf node that corresponds to this token. As an example, returning to the right side of \cref{fig:proposed:parser:trees}, the unexpected rule \mintinline[breaklines]{text}|'{G' -> '{' '{'| covers the first two tokens in the sentence. We consider this anomaly to be correctly localized because these tokens are adjacent to the token that was deleted.  We note that the unexpected rule \mintinline[breaklines]{text}|'{G}}' -> '{G}' '}'| covers the last token, and thus gives another candidate (though incorrect) location for an anomaly.  The rule \mintinline[breaklines]{text}|'{G}{G}' -> '{G}' '{G}'| does not cover any tokens, because it applies at a node that has no leaf nodes as children.

\subsection{Sentence simplification}\label{sec:proposed:simplification}

While the Simple-JSON format is highly structured, there are many formats where example sentences tend to contain a high degree of randomness/entropy.  For example, consider the {\em Key-List} language, where each sentence is a list of {\em keys}, such as \mintinline[breaklines]{text}|/cjc /i /sp|; each key begins with a forward slash, and the keys are separated by spaces.  The keys resemble those encountered in a {\em dictionary} within the PDF file format, but with randomly-generated content for each key.  Due to this randomness, parsing a key requires a large set of rules: e.g., \mintinline[breaklines]{text}|'/G' -> '/' 'a'|, \mintinline[breaklines]{text}|'/G' -> '/G' 'a'|, \mintinline[breaklines]{text}|'/G' -> '/' 'b'|, \mintinline[breaklines]{text}|'/G' -> '/G' 'b'|, ..., \mintinline[breaklines]{text}|'/G' -> '/' 'z'|, \mintinline[breaklines]{text}|'/G' -> '/G' 'z'|; here, a subgrammar merge rule, such as \mintinline[breaklines]{text}|'/G' -> '/' 'a'|, captures the requirement that a key must contain at least one letter after the forward slash; an anchored merge rule, such as \mintinline[breaklines]{text}|'/G' -> '/G' 'a'|, captures the possibility that the key might contain more than one letter.  It is non-trivial to learn such a set of rules; if the set of rules is not learned effectively, then a given new sentence may result in unexpected rules, and be labeled as anomalous.

To overcome this issue, we can attempt to differentiate between the low-entropy and high-entropy regions of an example sentence by applying our rule extraction algorithm, but filtering out production rules that occur infrequently within the parse trees from which they were extracted, treating such rules as unexpected (recall that $R$ is a multiset, where each rule is associated with the number of occurrences).  Then, we can use the remaining rules to perform anomaly localization as described previously, but with the goal of localizing high-entropy regions, rather than true anomalies.  Subsequently, we {\em simplify} each sentence in our datasets $(S, S^{\prime})$, by replacing each high-entropy region with the special {\em high-entropy token} \mintinline[breaklines]{text}|'&'|; then, a sentence such as \mintinline[breaklines]{text}|/cjc /i /sp| will be converted to \mintinline[breaklines]{text}|/& /& /&|.  Finally, we apply the pipeline of \cref{fig:proposed:parser:overview} a {\em second} time to the resulting simplified sentences $(\hat{S}, \hat{S^{\prime}})$, which should now be describable by a small set of rules that is easier to learn.

Unfortunately, for the {\em Key-List} language, the approach is problematic, because each low-entropy region consists of at most two tokens: a space and a forward slash.  Suppose that the second forward slash is removed in \mintinline[breaklines]{text}|/cjc /i /sp| to produce the anomalous sentence \mintinline[breaklines]{text}|/cjc i /sp|); then, the entire region \mintinline[breaklines]{text}|cjc i| will likely be covered by unexpected rules, treated as high-entropy, and collapsed to \mintinline[breaklines]{text}|'&'|; the simplified sentence will be \mintinline[breaklines]{text}|/& /&|, which is nominal; the anomaly will thus be missed.

To address this issue, in addition to the {\em topological} simplification approach, which uses the aforementioned notion of coverage by unexpected rules in the parse tree, we alternatively introduce {\em symbolic} simplification, where a token is said to belong to a high-entropy region if it is not found on the right side of any rule in $R$ (after infrequent production rules have been filtered out).  For the Key-List language, suppose that after filtering, $R$ consists of the rules \mintinline[breaklines]{text}|'/' -> ' ' '/'| and \mintinline[breaklines]{text}|'/' -> '/' '/'|; then, any symbol is considered to be part of a high-entropy region, unless it is a forward slash or a space token.  Under this symbolic approach, \mintinline[breaklines]{text}|/cjc i /sp| would be correctly simplified to \mintinline[breaklines]{text}|/& & /&|, which preserves the anomaly.  On the other hand, topological simplification can potentially be useful in situations where the low-entropy regions are larger, and where the same token may appear in both high-entropy and low-entropy regions, which is the case for many formats.


\begin{table*}[]
\centering
\caption{Anomaly detection and localization performance for Simple-JSON}
\label{table:results:simplejson}
{\footnotesize
\begin{tabular}{|l|c|c|c|}
\hline
\textbf{Anomaly}         & \textbf{True Positive Rate} & \textbf{Localization Rate} & \textbf{Localization Ratio} \\ \hline
Deleted Bracket & 94.7\% (100.0\%)              & 18.7\% (18.2\%)                 & 9.4\% (9.2\%)                      \\ \hline
Deleted Letter  & 100.0\% (100.0\%)              & 100.0\% (100.0\%)                 & 12.6\% (12.3\%)                      \\ \hline
Inserted Letter & 100.0\% (100.0\%)              & 61.0\% (62.1\%)                 & 7.4\% (7.5\%)                      \\ \hline
\end{tabular}

\bigskip
\begin{justify}
Each row corresponds to a particular anomaly, while columns capture performance metrics.  A given cell lists the average values of each metric without vs. with a validation set; specifically, the first value is obtained by averaging across all $30$ trials, while the second value (given in parentheses) is obtained by averaging only across those trials where the false positive rate was $0$ on the validation set; there were $18$ such trials.  For nominal evaluation sentences, the false positive rate was $6.9\%$ without a validation set, and $0\%$ with a validation set. The localization rate denotes the percentage of sentences where the anomaly is correctly localized (i.e., adjacent to or within the set of tokens that were labeled as anomalous by the algorithm); the localization ratio denotes the percentage of tokens that were labeled as anomalous, amongst the sentences where the anomaly was correctly localized.
\end{justify}
}
\end{table*}

\section{Evaluation}\label{sec:evaluation}

In our initial set of experiments, we applied our approach to randomly-generated sentences in the Simple-JSON grammar; the generation procedure is described in \cite{woods2021rlgrit}. We generated $120$ nominal sentences for training, $100$ nominal sentences for production rule extraction, $100$ nominal sentences for validation (explained in the caption to \cref{table:results:simplejson}) and $200$ sentences for the evaluation of anomaly detection, with $100$ of these sentences made anomalous in some way, as described below, and the rest nominal.  Training sentences were generated first, with duplicate sentences being permitted; the generated sentences were then randomly shuffled.  When generating the remaining sentences, duplicates were not permitted; the generated sentences were also randomly shuffled, before being split into sets for rule extraction, validation and evaluation, and before injecting anomalies into some of the evaluation sentences.  Since longer sentences are more likely to be unique, the $120$ training sentences tended to be shorter: $11.3$ tokens (characters), on average, compared with $27.0$ tokens for the extraction, validation, and nominal evaluation sentences (for anomalous evaluation sentences, the exact token counts depended on the type of anomaly).


The production rule extraction, anomaly detection, and anomaly localization algorithms are deterministic, though their results depend on the stochastically-trained RL-based parser.  To determine the effects of this stochasticity, we performed $30$ independent experimental trials, where in each trial, a parser was trained on the $120$ sentences, with a distinct set of randomly-generated neural network weights, a different permutation of the training set (during shuffling), and different random choices made during the RL process.  Training details are found in \cite{woods2021rlgrit}.  For a given trial, we applied the anomaly detection and localization procedures once to the $100$ nominal sentences in the evaluation dataset, and three times to the $100$ anomalous sentences, each time, with a different anomaly injected into the sentences: the first anomaly was the deletion of a single, randomly-chosen bracket (\mintinline[breaklines]{text}|{| or \mintinline[breaklines]{text}|}|); the second anomaly was the deletion of a single, randomly-chosen letter (\mintinline[breaklines]{text}|a|, \mintinline[breaklines]{text}|b|, or \mintinline[breaklines]{text}|c|); the final anomaly was the insertion of a single, randomly-chosen letter (\mintinline[breaklines]{text}|a|, \mintinline[breaklines]{text}|b|, or \mintinline[breaklines]{text}|c|) into some randomly-chosen location within the sentence.

To gauge the performance of anomaly detection, we captured the proportion (expressed as a percentage) of sentences that were labeled as anomalous; if the sentences were nominal, then this provides us with the {\em false positive} rate; otherwise, this is the {\em true positive rate}.  For the anomalous sentences, we additionally captured two anomaly localization metrics, which are both based on the set of tokens that the algorithm labels as potentially anomalous.  The first such metric is the {\em localization rate}, which is the proportion of sentences where this set of tokens contains the inserted token (for letter insertion anomalies), or a token that was immediately adjacent to the deleted token (for letter and bracket deletion anomalies).  The {\em localization ratio} is the proportion of tokens that were labeled as potentially anomalous, amongst only those sentences where localization was correct; for example, in the sentence \mintinline[breaklines]{text}|{a{b}}|, if only the token \mintinline[breaklines]{text}|a| is labeled as potentially anomalous, then the ratio is $\frac{1}{6}$; if all tokens are labeled as such, then the ratio is $1$.  A lower ratio indicates a more precise localization.

The results of the experiment are summarized in \cref{table:results:simplejson}.  For anomaly detection, the true and false positive rates show that using a validation set can potentially mitigate the variability that exists in the effectiveness of model training between different trials.  In a practical setting, we might discard a parser if its validation false positive rate is not sufficiently low, and attempt training again.  It is worth noting that even across the $18$ trials with perfect anomaly detection performance on the validation set, there was some variation in the rules that were learned from nominal sentences: for example, in some cases, the parser learned to perform right-biased merges, such as \mintinline[breaklines]{text}|'G}' -> 'a' '}'|, rather than left-biased merges, such as \mintinline[breaklines]{text}|'{G' -> '{' 'a'|.  Furthermore, instead of the anchored merge rule \mintinline[breaklines]{text}|'{G' -> '{G' '{G}'| (presented earlier in the paper), it was common to see either \mintinline[breaklines]{text}|'}' -> '{G}' '}'| or \mintinline[breaklines]{text}|'{' -> '{' {G}'| (with a bias that is opposite of the bias of other rules).


While our approach can thus be very effective for detecting several different types anomalies in the Simple-JSON format, the effectiveness of anomaly localization depends significantly upon the anomaly type, and is not necessarily improved by using a validation set.  Letter deletions are correctly localized in all cases; however, for bracket deletions and letter insertions, localization rates are much lower.  For letter insertions, our localization approach still outperforms a purely random strategy; such a strategy independently labels each token as potentially anomalous with probability $p$; this would result in an expected localization rate and ratio that are both $p$, while in our experiments, the localization rate is much higher than the localization ratio.  For bracket deletion anomalies, where there are typically two tokens adjacent to the deleted token (unless the deleted token was at the beginning or at the end of the sentence), the expected localization rate would have an upper bound of $2p - p^{2}$ (this is the probability that at least one of the adjacent tokens is labeled as anomalous), and the expected localization ratio would still be $p$.  With the random strategy, if $p = 9.2\%$, this would yield an expected localization ratio that matches the experimentally-observed localization ratio of $9.2\%$ in \cref{table:results:simplejson}, and an expected localization rate that is close to $2p - p^{2} = 2 \times 0.092 - 0.092^{2} = 17.6\%$, which is not far below the experimentally-observed value of $18.2\%$ (also in the table), so bracket deletions are not effectively localized.

For both bracket deletions and letter insertions, the localization problem can be inherently ambiguous; for example, given the anomalous sentence \mintinline[breaklines]{text}|{{{a}{b}}|, it is not possible to determine whether the missing right bracket was the sixth token in the uncorrupted sentence \mintinline[breaklines]{text}|{{{a}}{b}}| or the last token in \mintinline[breaklines]{text}|{{{a}{b}}}|; similarly, in \mintinline[breaklines]{text}|{ab}|, it is not clear whether \mintinline[breaklines]{text}|a| or \mintinline[breaklines]{text}|b| is the extra letter.  However, we have observed situations where unexpected rules do not cover any tokens at all, or only cover the tokens that could not possibly be anomalous.  An example of the former situation is shown in \cref{fig:proposed:parser:precedence}; here, the rule \mintinline[breaklines]{text}|'{G' -> '{G' '{G}'| has an unexpected precedence constraint, but it exists at a node that has no leaf nodes as children; thus, it does not cover any tokens.  We have experimented with an alternative notion of coverage, where a token is covered so long as its leaf node is a descendant (not necessarily a child) of a node with an unexpected rule or a rule with an unexpected precedence constraint, but found that this resulted in poorer localization performance, with large localization ratios (i.e., with many tokens labeled as potentially anomalous).  In \cref{sec:discussion}, we discuss a possible approach for improving localization accuracy.

To evaluate our simplification capability (\cref{sec:proposed:simplification}), we applied it to the Key-List dataset, where each sentence consisted of one to five keys, and each key consisted of a slash and one to three lowercase letter tokens, with the number of keys, the number of letters, and the choice for each letter drawn from uniform distributions; for the anomalous sentences, one randomly-chosen space or forward slash was deleted from a sentence.  The pipeline in \cref{fig:proposed:parser:overview} was applied in two passes.  We observed that during the first pass, the parser was unable to learn a set of rules that would precisely capture the Key-List language, due to the presence of high-entropy regions within the keys.  Nonetheless, it was possible to use this parser to simplify the sentences, following the symbolic approach, with any rule that appears fewer than $100$ times (in the parse trees from which $R$ was extracted) filtered from $R$; this would result in the rules \mintinline[breaklines]{text}|'/' -> ' ' '/'| and \mintinline[breaklines]{text}|'/' -> '/' '/'| being the only ones remaining.  Then, the pipeline was applied to the simplified sentences, in order to perform anomaly detection.  During training (on both the original and the simplified sentences), we increased the reward associated with anchored merges by a factor of $2$, relative to the Simple-JSON case (specifically, the $\alpha_{anchor}$ parameter in \cite{woods2021rlgrit} was set to $0.8$); this encouraged the learning of anchored merge rules such as \mintinline[breaklines]{text}|'/&' -> '/&' ' /&'|.  Apart from this change, procedures were similar to those used for Simple-JSON.  On the evaluation dataset (which also consisted of $100$ nominal and $100$ anomalous sentences), the false positive rate was $0\%$, even without a validation dataset; the true positive rate was $87\%$.  Missed detections occurred when the sentence consisted of just one key with the slash removed (e.g., \mintinline[breaklines]{text}|cjc|); such a sentence is simplified to a single token \mintinline[breaklines]{text}|'&'|, which was treated as nominal.

Finally, we performed experiments where we applied different variations of the two pass procedure to the {\em Simple-JSON-Stream} dataset, which was described in \cite{woods2021rlgrit}. This dataset consisted of sentences in the Simple-JSON format, but with a prefix and a suffix, each consisting of $5$ to $20$ random tokens, which could include all lowercase letters, \mintinline[breaklines]{text}|{| and \mintinline[breaklines]{text}|}|, with equal probability.  We attempted different variations on the procedure, e.g., with both symbolic and topological simplification, and with different rule filtering thresholds, but were not able obtain adequate anomaly detection results.  Topological simplification was able to perfectly identify the high-entropy prefixes and suffixes for some sentences; for example, \mintinline[breaklines]{text}|{hfsawpl{{a}}ygictfxk| was correctly simplified to \mintinline[breaklines]{text}|&{{a}}&|.  However, in many other cases, some simplification errors were present; e.g., \mintinline[breaklines]{text}|v{uptffaxlnnjh{{b}{b}{a}}plvalinjhxrmcjb| was simplified to \mintinline[breaklines]{text}|&{{&{b}{a}}&|, with the first \mintinline[breaklines]{text}|b}| region erroneously treated as high-entropy.  These errors occurred because the RL process, when applied to the original sentences, was somewhat sensitive to the high-entropy regions, and the rules that were learned for parsing the low-entropy regions of Simple-JSON-Stream were not the same as the ones that were learned for Simple-JSON, and did not capture these regions as effectively.  In turn, the simplification errors resulted in a set of simplified sentences that often deviated somewhat from the Simple-JSON format, with an underlying grammar that was more complex, and more difficult to infer.  As a result, the rule sets that were learned during the second pass were not sufficiently effective for anomaly detection. We postulate that future improvements to the RL algorithm would mitigate these issues.


\section{Discussion and Future Work}\label{sec:discussion}

While anomaly detection is a heavily researched problem, most work has focused on detecting anomalies in signals that are not modeled by production rules; for example, when detecting anomalies in the behavior of an aircraft, these signals may consist of real-valued variables and low-level switching events \cite{das2013mkad}.  Where production rules are an appropriate modeling paradigm, anomaly detection in unknown formats can, in theory, be solved if grammatical inference is solved. Given a grammar, we can generate a parser for that grammar, and then apply that parser to a sentence; if the parser fails, then the sentence can be labeled as anomalous.  However, it can sometimes be more natural to learn a parser, rather than a grammar, as was done in \cite{cowger2020icarus,woods2021rlgrit}.  As the resulting parser successfully produces a parse tree for any sentence, it cannot be used directly for anomaly detection.  A key contribution of this paper is that we have extended this approach to extract a grammar from the parser, and to use this grammar for detecting and localizing anomalies in unknown formats with data type recurrency or high-entropy regions.

The approach raises a key theoretical question: can our production rules (based on regular merges, anchored merges, and subgrammar merges), together with precedence constraints, describe any context-free language, or only a subset?  This question remains the subject of future work; specifically, we must formally analyze the extent to which the use of precedence constraints compensates for the fact that only a single subgrammar token \mintinline[breaklines]{text}|'G'| is used.  This allows for more effective learning, by decreasing the space of possible actions; however, it also results in a scarcity of nonterminal symbols: in our rule representation, each nonterminal symbol (e.g., \mintinline[breaklines]{text}|'{G'|) is a sequence of terminal symbols (tokens) and/or the subgrammar token \mintinline[breaklines]{text}|'G'|.  On the other hand, in a standard grammar formulation (e.g., the Chomsky Normal Form of \cite{chomsky1959normalform}), an unlimited number of nonterminal symbols can be used, and this eliminates the need for explicitly defining precedence constraints.  For example, in \cref{sec:proposed:enhancements} the rules \mintinline[breaklines]{text}|'{G' -> '{' 'a'| and \mintinline[breaklines]{text}|'{G' -> '{G' '{G}'| might be represented as \mintinline[breaklines]{text}|X -> '{' 'a'| and \mintinline[breaklines]{text}|Y -> Y Z|, respectively.  Then, there would be no concern about the output of the first rule being used as an input to the second rule, since \mintinline[breaklines]{text}|X|, \mintinline[breaklines]{text}|Y| and \mintinline[breaklines]{text}|Z| are distinct symbols.

On the empirical side, it is of interest to reevaluate the approach with a broader range of formats, as well as more complex anomalies (including anomalies that involve the insertion, deletion, or substitution of multiple tokens, rather than just a single token).  Another possible research direction is to improve the accuracy of anomaly localization, and furthermore, to provide automated suggestions on how a given anomaly could be corrected.  Here, we can potentially leverage machine learning approaches that have been developed for localizing and correcting syntax errors in programs written in languages such as Java \cite{santos2018sensibility} or Python \cite{bhatia2016synfix}.  Roughly speaking, these approaches train models that predict the probability of a given token in a program, given previous tokens; if the probability is low, then the token might be labeled as anomalous, and a suggested fix might involve replacing this token with a higher-probability token.  In some sense, these approaches may be viewed as complementary to our work, because they have been developed for programming languages with well-known formats, and rely on the presence of a compiler to determine if a particular suggested fix eliminates the anomaly; for unknown languages/formats, our anomaly detection procedure could potentially take on this role.

\section{Conclusions}\label{sec:conclusions}

In this paper, we presented an approach for extracting production rules from a neural network parser that was trained via the RL-GRIT algorithm, and for using extracted rules to detect and localize anomalies.  We demonstrated the effectiveness of the approach on datasets consisting of sentences in a non-regular (context-free) format (Simple-JSON) and a format that contains high-entropy regions (Key-List).  This suggests that with the extensions, the approach shows promise for performing anomaly detection in unknown formats.  At the same time, we have found that it is a challenge to apply the approach when high-entropy and low-entropy regions may contain some of the same tokens, as is the case in the Simple-JSON-Stream format.  It may be possible to mitigate this issue by further tuning the underlying RL algorithm, such that the presence of high-entropy regions has a less significant impact on the production rules that are learned for the low-entropy regions.  Our hope is that as these improvements are made, our approach will be sufficiently powerful to help better understand unknown formats. Since the RL-based approach is promising for the inference of non-trivial grammars, which underlie many real-world data formats, we believe this will prove useful for improving pre-filters aimed at making enterprise systems safer and more secure.

\section*{Acknowledgments}


This material is based upon work supported by the Defense Advanced Research Projects Agency (DARPA) under Contract No. HR0011-19-C-0073. Any opinions, findings and conclusions or recommendations expressed in this material are those of the author(s) and do not necessarily reflect the views of the Defense Advanced Research Projects Agency (DARPA).  The authors thank Richard Jones, Julien Vanegue, and anonymous reviewers for their feedback and advice in editing this work.


\bibliographystyle{./sty/ieee/IEEEtran-nomonth}
\bibliography{./sty/ieee/IEEEabrv,refs}

\end{document}